\documentclass[11pt,a4paper]{article}
\usepackage[hyperref]{acl2019}
\usepackage{times}
\usepackage{latexsym}
\usepackage{float}
\usepackage{url}
\usepackage{graphicx}
\usepackage{xcolor}
\usepackage{soul}

\aclfinalcopy 
\def\aclpaperid{1433} 

\title{Do Human Rationales Improve Machine Explanations?}

\author{Julia Strout, Ye Zhang and Raymond J. Mooney \\
  Department of Computer Science \\
  University of Texas at Austin \\
  \texttt{\{jstrout,yezhang,mooney\}@cs.utexas.edu} 
}

\date{}

\begin{document}
\maketitle
\begin{abstract}
Work on ``learning with rationales'' shows that humans providing explanations to a machine learning system can improve the system's predictive accuracy. However, this work has not been connected to work in ``explainable AI''  which concerns machines explaining their reasoning to humans. In this work, we show that learning with rationales can also improve the quality of the machine's explanations as evaluated by human judges. Specifically, we present experiments showing that, for CNN-based text classification, explanations generated using ``supervised attention'' are judged superior to explanations generated using normal unsupervised attention.
\end{abstract}

\section{Introduction}

Recently, the need for explainable artificial intelligence (XAI) has become a major concern due to the increased use of machine learning in automated decision making \cite{gunning2017explainable,ijcai17-xai-workshop}. 
On the other hand, work on ``learning with rationales'' \cite{zaidan2007using, zhang2016rationale} has shown that humans providing explanatory information supporting their supervised classification labels can improve the accuracy of machine learning. These human annotations that can explain  classification labels are called \textit{\textbf{rationales}}. In particular, for text categorization, humans select phrases or sentences from a document that most support their decision as rationales. 

However, there is no work connecting ``learning from rationales'' with improving XAI, although they are clearly complementary problems.  

\textbf{Contribution} We explore whether learning from human explanations actually improves a system's ability to explain its decisions to human users.  Specifically, we show that for explanations for text classification in the form of selected passages that best support a decision, training on human
rationales improves the quality of a system's explanations as judged by human evaluators.


Attention mechanisms \cite{bahdanau2014neural} have become standard practice in computer vision and text classification \cite{vaswani2017attention, yang2016hierarchical}. In both computer vision and text-based tasks, learned attention weights have been shown through human evaluation to be useful explanations for a model's decisions  \cite{park2018multimodal, rocktaschel2015reasoning, hermann2015teaching, xu2015show};  however, attention's explanatory power has come into question in recent work \cite{byron}, which we discuss in Section 2.


Traditional attention mechanisms are unsupervised; however, recent work has shown that supervising attention with human annotated rationales can improve learning for text classification based on Convolutional Neural Networks (CNNs) \cite{zhang2016rationale}. While this work alludes to improved explainability using supervised attention, it does not explicitly evaluate this claim. 
We extend this work by evaluating whether supervised attention using human rationales, rather than unsupervised attention, actually improves explanation. Explanations from both models are full sentences that the model has weighted as being most important to the document's final classification. 

While automated evaluations of explanations (e.g. comparing them to human gold-standard explanations \cite{lei2016rationalizing}) can be somewhat useful, we argue that because the goal of machine explanations is to help users, they should be directly evaluated by human judges. Machine explanations can be different from human ones, but still provide good justification for a decision \cite{das2017human}. This opinion is shared by other researchers in the area \cite{DoshiKim2017Interpretability}, but human evaluation is often avoided due to the time required and difficulty of conducting human trials. We believe it is a necessary element of explainability research, and in this work, we compare the explanations from the two models through human evaluation on Mechanical Turk and find that the model trained with human rationales is judged to generate explanations that better support its decisions.


\section{Related Work}
There is a growing body of research on explainable AI ~\cite{koh2017understanding,ribeiro2016should,li2016visualizing,hendricks2018grounding}, but it is not connected to work on learning with human rationales, which we review below.

As discussed above, \newcite{zhang2016rationale} demonstrate increased predictive accuracy of CNN models augmented with human rationales. Here, we first reproduce their predictive results, and then focus on extracting and evaluating explanations from the models.
 \newcite{lei2016rationalizing} present a model that extracts rationales for predictions without training on rationales. They compare their extracted rationales to human gold-standard ones through automated evaluations, i.e., precision and recall. 
\newcite{bao2018deriving} extend this work by learning a mapping from the human rationales to continuous attention. They transfer this mapping to low resource target domains as an auxiliary training signal to improve classification accuracy in the target domain. They compare their learned attention with human rationales by calculating their cosine distance to the `oracle' attention. 

None of the above related work asks human users to evaluate the generated explanations. However, \newcite{nguyen2018comparing} does compare human and automatic evaluations of explanations. That work finds that human evaluation is moderately, but statistically significantly, correlated with the automatic metrics. However, it does not evaluate any explanations based on attention, nor do the explanations make use of any extra human supervision.

As mentioned above, there has also been some recent criticism of using attention as explanation \cite{byron}, due to a lack of correlation between the attention weights and gradient based methods which are more ``faithful" to the model's reasoning. However, attention weights offer some insight into at least one point of internal representation in the model, and they also impact the training of the later features. Our contribution is to measure how useful these attention based explanations are to humans in understanding a model's decision as compared to a different model architecture that explicitly learns to predict which sentences make good explanations.

In this work, we have human judges evaluate both attention based machine explanations and machine explanations trained from human rationales, thus connecting learning from human explanations and machine explanations to humans.

\section{Models and Dataset}

\subsection{Models}
We replicate the work of \newcite{zhang2016rationale} and use a CNN as our underlying baseline model for document classification. To model a document, each sentence is encoded as a sentence vector using a CNN, and then the document vector is formed by summing over the sentence vectors. We use two variations of this baseline model, a rationale-augmented CNN (\textit{\textbf{RA-CNN}}) and an attention based CNN (\textit{\textbf{AT-CNN}}) \cite{yang2016hierarchical}. RA-CNN is trained on both the document label and the rationale labels. In this model, the document vector is a weighted sum of the composite sentence vectors, where the weight is the probability of the sentence being a rationale. In AT-CNN, the document vector is still a weighted sum of sentence CNN vectors, but the weight is not learned from rationales. Rather, a trainable context vector is introduced from scratch. We calculate the interaction between this context vector and each sentence vector to induce attention weights over the sentences. The only difference between RA-CNN and AT-CNN is that RA-CNN relies on the human annotated rationales to learn the sentence weight at training time, while AT-CNN learns the sentence weight without utilizing any human rationales. For the details of these two models and training see ~\newcite{zhang2016rationale}.


\subsection{Explanations}
At test time, each model can provide explanations for its classification decision by either choosing the sentences with the largest probability of being a rationale in RA-CNN or the sentences with the largest attention weights in AT-CNN. By comparing the quality of explanations output by the two models at test time, we can judge whether capitalizing on human explanations at training time can improve the machine explanations at test time. 

\subsection{Dataset}
We evaluate the explanations for both models on the movie review dataset from ~\newcite{zaidan2007using}. It contains 1,000 positive reviews and 1,000 negative reviews where 900 of each are annotated with human rationales. Each review is a document consisting of 32 sentences on average, and each annotated document contains about 8 rationale sentences. We use the 1,800 annotated documents as the training set, and the remaining 200 documents without extra annotation as test. The human rationales are used as supervision in RA-CNN but not in AT-CNN. 

\subsection{Classification Accuracy}
The classification accuracy of each model on the test set is summarized in Table \ref{table:model}. Since there is variance across multiple trials, we pick the best performing model across several trials for human evaluation of the explanations. 

Table \ref{table:model} reproduces \newcite{zhang2016rationale}'s finding that providing human explanations to machines at training time (RA-CNN) improves predictive accuracy compared to learning explanations without human annotations (AT-CNN). Our results differ slightly from theirs in that our AT-CNN also outperforms the baseline Doc-CNN. We attribute this difference to possible slight variations in our implementation of AT-CNN. 

Note there are other works on learning attention that could potentially increase the prediction accuracy~\cite{lin2017structured, devlin2018bert}, but none of them are directly comparable to RA-CNN. We introduced the smallest difference (whether the sentence vector is trained using the rationale label) between AT-CNN and RA-CNN to make a fair comparison between their generated explanations. 

The focus of this work is on evaluating explanations rather than predictive accuracy, so we turn our attention to the question: Does humans explaining themselves to machines improve machines explaining themselves to humans? We will explore this in the next section.

\section{Explanation Evaluation Methods}
We use Amazon Mechanical Turk (AMT) to evaluate the explanations from both AT-CNN and RA-CNN. 
\subsection{HIT Design} Our Human Intelligence Task (HIT) shows a worker two copies of a test document along with the document's classification. Each copy of the document has a subset of sentences highlighted as explanations for the final classification. This subset is chosen as the 3 sentences with the largest weights from either AT-CNN's attention weights or RA-CNN's supervised weighting. We also evaluated a baseline model that selects 3 sentences at random. Given two randomly ordered documents, a worker must choose which document's highlighted sentences best support the overall classification. If the worker determines that both are equally supportive (or not supportive), then they can select `equal'. We only show workers documents that were correctly classified by both models. This resulted in 166 documents from the 200 in the test set. An example from one HIT is in Appendix A.

\begin{table}
\centering
\begin{tabular}{ c c c }
 Doc-CNN & AT-CNN & RA-CNN \\ 
 \hline
 86.00\% & 88.50\% & 90.00\% \\
\end{tabular}
\caption{Classification accuracy for movie reviews.}
\label{table:model}
\vspace{-10pt}
\end{table}

\begin{table*}[]
    \footnotesize
    \centering
    \begin{tabular}{|p{.7cm}|p{.6cm}|p{6.5cm}|p{6.5cm}|}
        \hline
        \bf{Label} & \bf{Rank} & \bf{AT-CNN} & \bf{RA-CNN} \\
        \hline
         & \center 1& archer is also bound by the limits of new york society , which is as intrusive as any other in the world. & the performances are absolutely breathtaking.\\
        \cline{2-4}
      
        \center Pos & \center 2 & the marriage is one which will unite two very prestigious families , in a society where nothing is more important than the opinions of others . & there are a few deft touches of filmmaking that are simply outstanding , and joanne woodward' narration is exquisite. \\
        \cline{2-4}
       
         & \center 3 & the supporting cast is also wonderful , with several characters so singular that they are indelible in one's memory . & the supporting cast is also wonderful , with several characters so singular that they are indelible in one's memory .
 \\
         \hline
         
         & \center 1 & soon the three guys are dealing dope to raise funds , while avoiding the cops and rival dealer sampson simpson (clarence williams iii) . & it's just that the comic setups are obvious and the payoffs nearly all fall flat .\\
         \cline{2-4}
          \center Neg & \center 2 & only williams stands out (while still performing on the level of his humor-free comedy rocket man) , but that is because he's imprisoned throughout most of the film , giving a much needed change of pace (but mostly swapping one set of obvious gags for another) . & watching the film clean and sober , you are bound to recognize how truly awful it is .\\
          \cline{2-4}
           & \center 3 &watching the film clean and sober , you are bound to recognize how truly awful it is . & the film would have been better off by sticking with the `` rebel'' tone it so eagerly tries to claim. \\
           \hline
    
    \end{tabular}
    \caption{Top 3 explanations from both models for both a positive and negative correctly classified test document. }
    \label{tab:examples}
\end{table*}

\subsection{Quality Control}
 In an effort to receive quality results from the crowd, we employ two strategies from crowd-sourcing research: gold standard questions and majority voting \cite{Hsueh:2009:DQC:1564131.1564137, eickhoff2013increasing}. Gold standard questions are designed to weed out unreliable workers who either do not understand the goal of the task or are poor workers. If a worker gets the gold standard question wrong, then we assume that their other responses are untrustworthy and do not use them. 

We also employ majority voting, which requires that at least two workers who pass the gold standard question agree on an answer. For greater than 90\% of the test documents, a majority vote was found after having three workers perform the task. Less than 10\% of the test documents required a fourth worker who passed the gold question to break a tie. We also chose to require the `Master' qualification that AMT uses to designate the best workers on the platform.
\begin{table}[]
    \centering
    \begin{tabular}{c|c|c}
         RA-CNN & AT-CNN & Equal  \\
         \hline
         43.47\% & 20.48\% & 36.14\% \\ 
    \end{tabular}
    \caption{AMT  results  comparing  explanations from  RA-CNN  to  AT-CNN.  Workers  were  asked  to choose which document's highlighted sentences were a better explanation for the final classification.}
    \label{tab:results}
\end{table}

\begin{table}[]
    \centering
    \begin{tabular}{c|c|c}
         AT-CNN & Random & Equal  \\
         \hline
         57.23\%& 15.66\%& 27.12\% \\
    \end{tabular}
    \caption{AMT results comparing AT-CNN to the random baseline.}
    \label{tab:random}
    \vspace{-7pt}
\end{table}

\section{Explanation Evaluation Results}

Table \ref{tab:results} contains the results for comparing the top 3 explanations from AT-CNN to the top 3 explanations from RA-CNN for the 166 test documents where the models each correctly classified the document. The statistics presented are the percentage of times reliable workers agreed that one model's explanations better supported the document's classification or were equal.

Overall, it is clear that RA-CNN is providing better explanations for the plurality of test documents (43.47\%). The explanations are considered equal 36.14\% of the time, and the remaining 20.48\% of the documents were better explained by AT-CNN.

After seeing these results, we decided to run another baseline test to ensure that AT-CNN explanations are reasonable and can at least beat a weak baseline. The results from comparing AT-CNN explanations to randomly sampled sentences from the test document are in Table \ref{tab:random}. From these results we can see that AT-CNN is beating the random baseline the majority of the time, demonstrating that attention, even without human supervision, can provide helpful explanations for a model's decision.

To further understand the differences between the explanations from AT-CNN and RA-CNN, we calculated statistics to find the amount of overlap in the top three explanatory sentences from each model. In 33.5\% of the test documents, the models share no explanation sentences, in 43.1\% they share one explanation sentence, in 22.2\% they share two explanation sentences, and they share all three in 1.2\%. When considering just the most highly weighted sentence, or top explanation, the models agree 18.6 \% of the time. So while it is relatively rare for the models to produce the same top explanatory sentence, we chose to show humans three explanatory sentences per test document to provide insight even in those matching cases.

Table \ref{tab:examples} contains the top 3 explanations from each model for two test documents. In both examples, AT-CNN extracts sentences that are more plot related and give less insight into the reviewer's opinion as compared to RA-CNN. These sentences are generally less helpful for understanding the classification of the movie review. In the second example, both models have identified a good explanatory sentence: ``watching the film clean and sober, you are bound to recognize how truly awful it is.'' However, AT-CNN ranks it as less important than two sentences that primarily describe the plot of the film while RA-CNN only ranks another, equally explanatory sentence as more important. 

An interesting future avenue for evaluation is to compare explanations from when the models make incorrect predictions. We found a trend in the explanations for test documents that both models misclassified where RA-CNN produced explanations that supported the misclassification while AT-CNN produced more explanations that supported the correct classification, despite the model's decision. While this analysis is too small scale to be conclusive, this raises the question for future work: Do we want our explanation systems to offer the best support for the chosen decision or would it be more beneficial if they provide an explanation that brings the decision into question? 

\section{Conclusion}
This paper has demonstrated that training with human rationales improves explanations for a model's classification decisions as evaluated by human judges. We show that while an unsupervised attention based model does provide some valuable explanations, as proven in the experiments comparing to a random baseline, a supervised attention model that trains on human rationales outperforms those results. 


\section*{Acknowledgements}
This research was supported by the DARPA XAI
program through a grant from AFRL.
The views
and conclusions contained herein are those of the authors
and should not be interpreted as necessarily representing
the official policies or endorsements, either expressed or implied,
of the U.S. Government.
\bibliography{acl2019}
\bibliographystyle{acl_natbib}
\clearpage
\appendix
\section{Sample HIT}
\begin{figure}[h!]
\centering
\onecolumn\includegraphics[width=15cm]{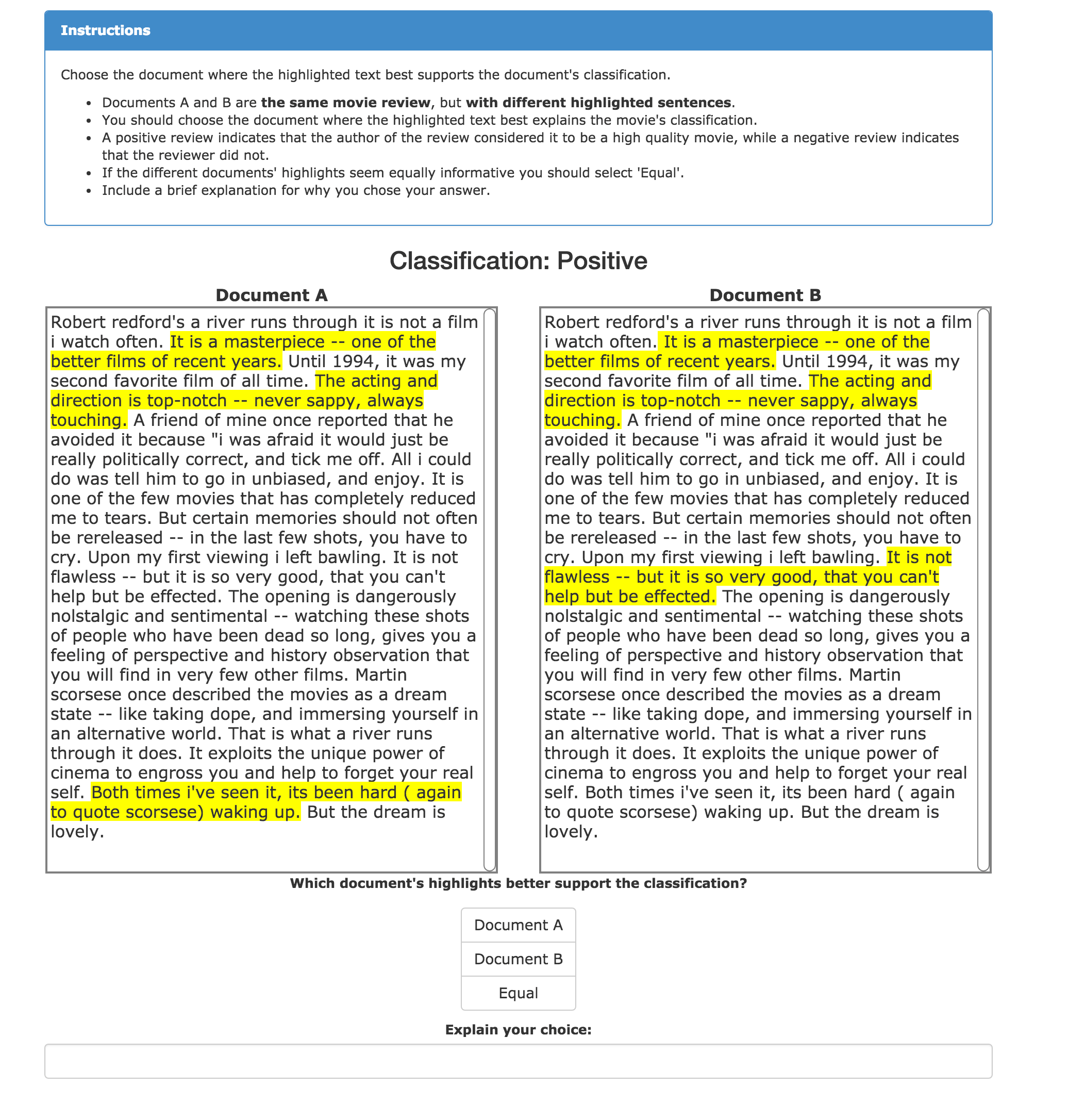}
\caption{A sample HIT asking workers to compare two explanations for the same movie review.}
\end{figure}
\end{document}